\documentclass[conference]{IEEEtran}
\IEEEoverridecommandlockouts
\usepackage{cite}
\usepackage{amsmath,amssymb,amsfonts}
\usepackage{algorithmic}
\usepackage{graphicx}
\usepackage{textcomp}
\usepackage{xcolor}
\usepackage{hyperref}
\usepackage{placeins}
\usepackage{booktabs}
\usepackage{graphicx}
\usepackage{caption}
\usepackage{subcaption}
\usepackage{multirow, multicol}

\def\BibTeX{{\rm B\kern-.05em{\sc i\kern-.025em b}\kern-.08em
    T\kern-.1667em\lower.7ex\hbox{E}\kern-.125emX}}
\begin{document}

\title{YouLeQD: Decoding the Cognitive Complexity of Questions and Engagement in Online Educational Videos from Learners' Perspectives}

\author{\IEEEauthorblockN{Anonymous Submission}
\IEEEauthorblockA{}
}

\newcommand{\minus}{\scalebox{0.75}[1.0]{$-$}}

 \author{\IEEEauthorblockN{Nong Ming}
\IEEEauthorblockA{\textit{Department of Computer Science } \\
 \textit{ Kennesaw State University}\\
 Kennesaw, USA\\
 nming@students.kennesaw.edu}
 \and
 \IEEEauthorblockN{Sachin Sharma}
 \IEEEauthorblockA{\textit{Department of Computer Science} \\
 \textit{Kennesaw State University}\\
 Kennesaw, USA \\
 ssharm21@students.kennesaw.edu}
 \and
 \IEEEauthorblockN{Jiho Noh}
 \IEEEauthorblockA{\textit{Department of Computer Science} \\
 \textit{Kennesaw State University}\\
 Kennesaw, USA \\
 jnoh3@kennesaw.edu}
 }
\maketitle

\begin{abstract}

Questioning is a fundamental aspect of education, as it helps assess students' understanding, promotes critical thinking, and encourages active engagement. With the rise of artificial intelligence in education, there is a growing interest in developing intelligent systems that can automatically generate and answer questions and facilitate interactions in both virtual and in-person education settings. However, to develop effective AI models for education, it is essential to have a fundamental understanding of questioning. In this study, we created the YouTube Learners' Questions on Bloom's Taxonomy Dataset (YouLeQD), which contains learner-posed questions from YouTube lecture video comments. Along with the dataset, we developed two RoBERTa-based classification models leveraging Large Language Models to detect questions and analyze their cognitive complexity using Bloom's Taxonomy. This dataset and our findings provide valuable insights into the cognitive complexity of learner-posed questions in educational videos and their relationship with interaction metrics. This can aid in the development of more effective AI models for education and improve the overall learning experience for students. 

\end{abstract}


\section{Introduction}

Questioning is a fundamental aspect of education, as it plays a crucial role in learning. Instructors use questions to assess students' understanding, promote critical thinking, and encourage active engagement in the classroom. An effective question can initiate a thought-provoking discussion, stimulate curiosity, and enhance comprehension among peer students, resulting in improved learning performance in educational settings~\cite{king1990enhancing,chin2002student,rosenshine1996teaching}. In recent years, there has been a growing interest in developing generative artificial intelligence (GenAI) models and conversational agents for educational purposes. These models aim to provide personalized learning experiences~\cite{borah2024improved}, answer students' questions~\cite{nyaaba2024generative}, and facilitate interactions between students and instructors~\cite{luo2024does}. In the pursuit of creating intelligent systems for education, it is essential to have a comprehensive understanding of the questioning practice.

In the field of education, the study of questioning has been an active research topic for a long time. Various studies have examined the types of questions posed by students and instructors, the cognitive complexity of these questions~\cite{grainger2018process ,teplitski2018student}, and the impact of questioning strategies on learning outcomes~\cite{irons2014coaching}. A diverse range of questioning strategies have been devised, incorporating instructional designs such as Gagne's nine events of instruction~\cite{kruse2009gagne}, Kolb's experiential learning~\cite{bergsteiner2010kolb}, and Merrill's first principles of instruction~\cite{merrill2002first}. Furthermore, researchers have developed question-answering systems, chatbots, and educational tools to foster student engagement and enhance learning experiences~\cite{kuhail2023interacting,lee2022developing}.

The application of GenAI models in the field of education shows significant potential, particularly in the automated generation of questions for pedagogical purposes. Recently, the development of Large Language Models (LLMs) has significantly strengthened Automatic Question Generation (AQG) capabilities. Studies have shown that LLMs are capable of producing high-quality questions with clear language for various subjects~\cite{tran2023generating} and generating questions aligned with Bloom's Taxonomy levels to determine the learner's cognitive level~\cite{gnanasekaran2021automatic}. The efficacy of AI-generated questions has been assessed through both manual evaluation by subject matter experts and automated evaluation by other LLMs, as demonstrated in previous research~\cite{van2022evaluating,bhat2022towards}. 


In the field of Artificial Intelligence (AI), however, the applications of questioning strategies often lack a comprehensive grasp of instructional design principles. This can lead to ineffective or even harmful use of AI in educational environments. As such, it is crucial to bridge the gap between AI and instructional design, in order to ensure that AI-driven questioning strategies are thoughtfully crafted and executed to facilitate effective learning outcomes.

This research aims to explore the intersection of AI and pedagogical strategies in the context of questioning. By examining questions from online educational videos and their alignment with instructional design principles, we have developed the \textit{YouTube Learners' Question Dataset (YouLeQD)}, which consists of transcripts and learner-posed questions extracted from comments of 1,762 lecture videos. Although YouTube is not primarily an educational platform, it has become a popular source of educational content and an interface for learners' engagement through the comments section. We collected transcripts and comments from lecture videos in five STEM subjects (i.e., biology, computer science, mathematics, chemistry, and physics) and extracted 57,242 questions. We analyzed these questions based on their cognitive levels of learning outcomes using Bloom's taxonomy. Our study delves further into the relationship between the cognitive levels of questions and the engagement levels of comments. The following questions are the main focus of our study:

\begin{itemize}
  \item RQ1: How do the learner-posed questions in the YouTube educational video comments align with Bloom's Taxonomy cognitive complexity levels?
  \item RQ2: Is there a relationship between the cognitive levels of questions and the interaction rates, specifically the number of likes and replies to the comments? 
  \item RQ3: Are there differences in question type distributions by cognitive complexity across five subjects?
\end{itemize}


The findings of this study will provide valuable insights into the cognitive complexity of learner-posed questions in online educational videos and their relationship with engagement metrics. This research will contribute to the development of effective AI models for education. 

The main contributions of this study are as follows: 
\begin{enumerate}
  \item We introduce a publicly accessible dataset containing transcripts from YouTube educational videos, along with questions posted by online learners.
  \item We propose question detection and classification methods based on Bloom's Taxonomy levels using fine-tuned RoBERTa models, leveraging large language models in the learning process.
  \item We provide an in-depth analysis of the cognitive complexity of learner-posed questions and their relationship with interaction rates.
\end{enumerate}

The YouLeQD dataset and code of the classification models are publicly available to support further research in this area~\footnote{https://github.com/Jiho-YesNLP/QYTL}.

\section{Related Work}

\subsection{YouTube as a Source of Educational Content}

YouTube offers a diverse range of educational materials, including highly regarded lecture videos from channels such as \textit{YaleCourses} and \textit{MIT OpenCourseWare}. Studies have demonstrated the efficacy of YouTube videos in various educational contexts, such as patient education~\cite{jamleh2021evaluation}, nurse training~\cite{clifton2011can}, and language learning~\cite{wahyuni2021use}. Furthermore, YouTube has been shown to be a valuable tool for educators, as it effectively captures students' attention and promotes their creativity~\cite{fadhil2020investigating}.

The use of YouTube's communication features, such as comments and like/dislike buttons, has been examined for its potential implementation in the field of education. Previous studies have shown that YouTube comments can support collaborative knowledge construction and deeper learning in science~\cite{dubovi2020empirical}, provide insights into how students engage with informational videos for informal learning~\cite{meyers2014comment}, and empower learners to engage in the content through comments~\cite{lee2017making}. These works have demonstrated the great potential of YouTube comments for pedagogical purposes. However, they mainly focus on how YouTube videos and comments can enhance learners' experience and encourage them to engage in the learning process. Only a few studies have explored the potential of YouTube video comments as a valuable source of information~\cite{wang2023sight} for incorporating recent GenAI technologies. The objective of this study is to investigate the potential of materials in online lecture videos to understand questioning activities in education and how they lead to constructive engagement and interaction among learners.

\subsection{Questioning in Education and Evaluation Methods}

Questioning is an essential way of engaging with students and promoting critical thinking in the classroom, whether it's an assessment question from the teacher or a question from a student in discussion~\cite{papinczak2012using,teplitski2018student}. Dillon~\cite{dillon2004questioning} discusses the important role that questioning plays in both formal and informal educative processes, emphasizing the need for teachers to ask open-ended questions that encourage students to think critically and creatively in order to better understand the subject.


In the field of education, significant efforts have been made to establish quality standards for evaluating questions. Among them, metric-based approaches assess the linguistic quality of questions, encompassing factors such as readability, relevance, grammar, fluency, and clarity~\cite{gatt2018survey,amidei2018evaluation,kurdi2020systematic,mulla2023automatic}. Post-hoc methods have been utilized to evaluate questions based on their difficulty~\cite{haladyna2021using}, discrimination~\cite{gierl2016evaluating,wind2019exploring}, and distractor effectiveness~\cite{vsmida2024developing,rezigalla2024item} drawing upon student response data. However, these approaches have limitations in their ability to capture the cognitive complexity of the questions and their potential to foster higher-order thinking skills. 

To address these limitations, recent research has proposed the utilization of cognitive complexity models as a means of evaluating the quality of questions. Prominent cognitive models, such as Bloom's taxonomy~\cite{bloom1956handbook}, six facets in Understanding by Design~\cite{roth2007understanding}, Marzano and Kendall's cognitive system~\cite{marzano1998implementing}, and Dee Fink's Significant Learning model~\cite{fink2013creating}, offer a comprehensive framework for categorizing questions based on their cognitive complexity. This allows educators to effectively assess the extent to which questions promote higher-order thinking skills. For instance, Bloom's taxonomy classifies questions into six levels of cognitive complexity: (i) knowledge, (ii) comprehension, (iii) application, (iv) analysis, (v) evaluation, and (vi) synthesis, with higher levels requiring more advanced thinking skills. Table \ref{tab:BT-levels} lists the cognitive levels of Bloom's taxonomy and provides examples illustrative examples of questions for each level.

\begin{table}[htbp]
  \centering
  \caption{Learning Objectives in Bloom's Taxonomy of Cognitive Levels}
  \label{tab:BT-levels}
  \begin{tabular}{@{}lp{0.75\linewidth}@{}}
    \toprule
    \textbf{Cognitive Level} & \textbf{Description and Example} \\
    \midrule
    Knowledge & 
      \begin{tabular}{@{}l@{}}
        Recall of information.\\
        \textit{(List the planets in the solar system.)}
      \end{tabular} \\ \midrule
    Comprehension & 
        \begin{tabular}{@{}l@{}}
          Understanding Concepts.\\
          \textit{(Explain the water cycle.)}
        \end{tabular} \\ \midrule
    Application& 
      \begin{tabular}{@{}l@{}}
        Applying knowledge in different contexts.\\
        \textit{(Use the quadratic formula to solve a given } \\
        \textit{  quadratic equation.)}
      \end{tabular} \\ \midrule
    Analysis &
      \begin{tabular}{@{}l@{}}
        Breaking down information, Identifying \\relationships and organization. \\
        \textit{(Compare four ways of serving foods made with} \\
        \textit{  seafood and examine which ones have the highest} \\
        \textit{  health benefits.)}
    \end{tabular} \\ \midrule
    Evaluation&
      \begin{tabular}{@{}l@{}}
        Judging and critiquing based on established criteria.\\
        \textit{(Which kinds of apples are suitable for baking a pie,} \\
        \textit{  and why?)}
      \end{tabular} \\ \midrule
    Synthesis &
      \begin{tabular}{@{}l@{}}
        Creating new ideas or solutions.\\
        \textit{(Make a business plan for a local company.)}
      \end{tabular} \\
    \bottomrule
  \end{tabular}
\end{table}

In the context of AQG system development and evaluation of generated questions, BT can guide question generation that targets specific levels of cognitive skills and knowledge, and assess the quality and complexity of the questions. Notably, Mohammed and Omar~\cite{mohammed2020question} have devised a straightforward question classification model based on BT, utilizing modified TF-IDF and Word2vec embeddings. Similarly, Ahmed and Sorour~\cite{ahmed2024classification} have developed an intelligent system for evaluating higher education exam papers by categorizing questions according to BT levels. Das et al.~\cite{das2020identification} have proposed two methods, using LDA and BERT respectively, for automatically identifying the BT level of assessment questions in various college-level subjects. In line with this ongoing research, our study aims to contribute to the field by proposing a method of using a cognitive model to evaluate the quality of questions in discussion forums, specifically within the context of YouTube comments. We aim to closely examine the characteristics of questions in online lecture platforms, providing valuable insights for researchers.

\subsection{LLM as Automatic Annotators}

In recent years, Large Language Models (LLMs) have demonstrated remarkable performance in various natural language processing (NLP) tasks, greatly revolutionizing the field~\cite{fan2023bibliometric}.  They also demonstrate great capability as automatic annotators for large-scale NLP tasks. Previous studies have shown that LLMs can serve as effective alternatives to non-expert crowdworkers~\cite{aguda2024large} for annotating financial data.  Studies on the potential capabilities of LLMs have shown that LLMs could efficiently provide explanations for labels as few-shot prompts for better labeling~\cite{he2023annollm}. Also, LLMs can use labeled samples from the current dataset as guidance to produce consistent labels~\cite{chen2024large}. In the educational domain, Rose Wang et al.~\cite{wang2023sight} employed GPT-4 as a scaling annotator to classify over 15,000 educational video comments as student feedback using their own rubric. Note that their primary focus was on analyzing the feedback itself, rather than synthesizing it for instructors for potential pedagogical use.


In light of these findings, our study utilizes LLMs to annotate questions while training classification models via data augmentation and knowledge distillation methods to accurately detect and classify these questions. The findings of our work will provide valuable insights into question generation in the educational field. 
 
\section{Methodology}

This section delves into the development of YouLeQD, outlining our data collection procedures, which involve the acquisition of transcripts and comments and question extraction. In the later section, we will detail the question classification models and the training methods.

\subsection{Data Acquisition: Transcripts and Comments} 

\begin{table*}[htbp]
    \centering
    \caption{Explicit Statistics of YoulQD. (Each metric is averaged across all videos in the subject and length is in tokens.)}
    \label{tab:qext-stats}
    \begin{tabular}{@{}lrrrrrrr@{}}
    \toprule
    Subject & \# Videos & \# Views & \# Likes &  \# Comments &
         \begin{tabular}{@{}r@{}}Transcript \\length \end{tabular} &
         \begin{tabular}{@{}r@{}}Comment \\length \end{tabular} &
         \begin{tabular}{@{}r@{}}\# Extracted \\questions \end{tabular} \\
    \midrule
         Biology          &  368 &  782K &  9.0K &  560 & 6,010  & 22 & 16,024\\
         Chemistry        &  356 &   37K &  0.5K &  33 &  3,544 & 18 & 2,050\\
         Mathematics      &  379 &  327K &  5.1K &  186 &  3,047 & 20 & 9,589\\
         Physics          &  432 &  533K &  7.5K &  255 &  3,256 & 17 & 11,354\\
         Computer Science &  352 &  325K &  5.6K &  151 &  2,897 & 18 & 18,789\\
    \bottomrule
    \end{tabular}
\end{table*}

Similar to previous studies~\cite{wang2023sight,chen2024qgen}, we collected educational lecture videos from YouTube, focusing on five STEM subjects: biology, computer science, mathematics, chemistry, and physics. The videos were selected from popular educational channels and playlists, including MIT OpenCourseWare (MIT 7.016, MIT 8.421, MIT 18.06), Stanford Online Course (CS229), and other professional educational channels. We tried to ensure that each subject was represented by a minimum of 300 videos, with a balanced distribution of videos across all subjects. In total, we collected 1,762 educational videos from YouTube, with an average of 352 videos per subject.


We used the Google Data API and publicly available Python libraries to download the video data, including the video ID, title, description, transcripts, comments, and metadata such as views and likes. The statistics of the dataset are shown in Table \ref{tab:qext-stats}. We limited the number of comments per video to 1,000 to avoid bias towards popular videos. We also removed comments that are too short or too long, as they are likely irrelevant or not informative.

\subsection{Question Extraction from Comments}

The classification of sentences based on their function, such as declarative, interrogative, and imperative, is a fundamental task in NLP, and it is widely considered a successfully solved problem. Among the previously proposed methods, the Transformer-based models have shown remarkable performance in sentence classification tasks. In this research, we train a neural network model to detect and extract question sentences from the collected comments. We fine-tune a RoBERTa model~\cite{liu2019roberta} on a publicly available dataset for interrogative sentence classification~\cite{Khan_2021}, which is built upon other datasets such as SQuAD~\cite{rajpurkar2016squad} and SPAADIA~\cite{leech2013spaadia}. The dataset comprises 211,168 sentences, including 80,167 non-interrogative sentences and 131,001 interrogative sentences, each with binary labels. The author of the dataset has deliberately removed question marks from some of the examples in order to prevent the model from overfitting to the punctuation marks.

To further improve the RoBERTa model, we adopted the Knowledge Distillation (KD) learning framework, which involves training a smaller student model to mimic the behavior of a larger teacher model~\cite{hinton2015distilling} using soft labels. In our case, we used GPT-4o as the teacher model to train the RoBERTa classification model. We asked GPT-4o to classify sentences into either interrogative or non-interrogative categories and utilized the probability distribution of the predictions in the optimization process. The samples with low log probabilities show difficult examples where confusion may arise (See examples in Appendix~\ref{appx:low-conf-sample}). For optimization, we formed the loss function inspired by the work of Zhou et al.~\cite{zhou2021rethinking}, combining typical cross-entropy (CE) loss and KL divergence loss between the probability distribution of the teacher and the one of student model as below: 

\begin{equation} \label{eq:kd}
  \mathcal{L} = \mathcal{L}_{CE} + \alpha \cdot \mathcal{L}_{KL},
\end{equation}

\begin{align*}
  \mathcal{L}_{CE} &= -\log \hat{y}_{i}^s, \\
  \mathcal{L}_{KL} &= \left( 1 - \exp \left( - \frac{\log \hat{y}_{i}^s}{\log \hat{y}_{i}^t} \right) \right) \cdot \left( - \tau^2 \sum_k \hat{y}_{k,\tau}^t \log \hat{y}_{k,\tau}^s \right),
\end{align*}

where $i$ is the target label and $\tau$ is the temperature parameter. Following the original work, we set $\tau=2$ for the probability smoothing and $\alpha=2.5$ for the soft loss weight. Additionally, in the context of binary classification, we make the assumption that $\hat{y}_0^t = 1- \hat{y}_1^t$.

After training, we use the fine-tuned RoBERTa model to classify the comments and extract the questions. We removed emojis, special characters (e.g., @ mentions), and URLs from the comments. We filter out the questions that are too short or too long (less than 3 tokens or more than 50 tokens) to ensure the quality of the extracted questions. The extracted questions are then stored in a dataset for further analysis.

\subsection{Bloom's Taxonomy Classification Model for Questions}

Once questions are extracted from the comments, we classify them based on their cognitive complexity using Bloom's Taxonomy (BT) levels. To achieve this, we fine-tune another RoBERTa model on a publicly available question classification dataset~\cite{das2020identification}, hereinafter referred to as DASQBT, which comprises 2,533 educational questions with the labels in BT cognitive level. 80\% of the examples are used for training and validation, while the remaining 20\% are reserved for evaluating the model's overall performance and generalization capabilities.

We trained the model using the Adam optimizer with a learning rate of \texttt{1e-05}, a batch size of 16, and a maximum sequence length of 128. The final two additional dense layers are trained with a dropout rate of 0.2 and ReLU activation for the purpose of classification.

\subsubsection{Data augmentation}

One of the challenges of utilizing DASQBT is the relatively small size of the dataset, which may limit the model's ability to generalize effectively. To obtain more examples, we employed a Large Language Model, specifically GPT-4o, to generate questions with cognitive-level labels. The expanded dataset now comprises 5,779 examples. The results presented in Table~\ref{tab:gpt-labels-agreement} demonstrate a high level of consistency between the labels generated by GPT and the original labels

\begin{table}[hbt]
    \centering
    \caption{Agreement between GPT-Predicted Labels and Original Question Classification Labels}
    \label{tab:gpt-labels-agreement}
    \begin{tabular}{@{}lrrrr@{}}
    \toprule
    \textbf{Model} & \textbf{Accuracy} & \textbf{Precision} & \textbf{Recall} & \textbf{F1-score}\\
    \midrule
    GPT-4o & 0.988 & 0.998 & 0.982 & 0.989 \\
    \bottomrule
    \end{tabular}
\end{table}

Appendix~\ref{appx:prompts} outlines the prompt utilized for the data augmentation. This prompt first lists the predominant action verbs associated with each BT cognitive level and a few examples of question generation. The prompt asks explicitly for generating an educational question of a designated type (e.g., Comprehension) on a selected topic (e.g., gene regulation) within a chosen subject (e.g., biology) where the topic is randomly selected from a manually curated pool of topics for each subject.

\subsubsection{Detecting Out-of-Distribution Examples and Evaluation Strategy}

Another challenge in training a BT classification model is the existence of Out-of-Distribution (OOD) examples within the extracted questions from YouTube comments; for example, irrelevant or nonsensical questions that cannot be categorized into any of the BT classes. The particular training dataset we used in this study contains examples that are highly relevant to the educational context and representative of the BT cognitive levels. However, the extracted questions from YouTube comments are mostly low-quality and irrelevant, which may not align with the BT cognitive levels (see Appendix~\ref{appx:question-type-review}). Classification models struggle to effectively generalize when the training and test data distributions differ significantly.

To address this issue, we conducted a two-stage training for the BT classification. In the first stage, we trained the model on the original DASQBT dataset. Then, we used the optimized model to predict BT labels with the questions and obtain the softmax probability distribution along with the predictions. Hendrycks and Gimpel stated that softmax probabilities are not directly applicable as confidence estimates. Still, they can effectively determine whether an example is misclassified or from a different distribution than the training data~\cite{hendrycks2022baseline}. This study uses the probabilities to detect irrelevant (or out-of-distribution) questions. We identified the top 500 examples with low maximum/predicted class probability from the softmax distributions and labeled them as belonging to the `Irrelevant' class.

In the following stage, we trained the same type of model on the expanded dataset, which includes additional question samples classified as `Irrelevant,' and evaluated its performance on the reserved test dataset. To further evaluate the accuracy and robustness of this model, we also tested the model on a human-labeled question sample. This human-annotated sample consists of 300 randomly selected questions, with an equal distribution of 60 questions per subject. The labeling process is carried out by three graduate-level students who are thoroughly briefed on the particular BT classification task. We take the most voted label as the final label for each question. This sample serves as an independent external test set to assess the model's performance.

\section{Experiments and Results}

\subsection{Question Extraction}

We trained a more robust RoBERTa model for question classification with the utilization of Knowledge Distillation (KD). The model achieved an f1-score of 99.42\% on the test set, which is a 1.31\% improvement over the initial model trained without KD. The results demonstrate that incorporating the KD method into the training process can enhance the performance of the RoBERTa model for question classification. It is worth noting, however, that the performance of the initial model was already high, which limited the potential for further improvement.

\subsection{Alignment of Questions with Bloom's Taxonomy}
We conducted experiments to evaluate the performance of the BT classification models to investigate the effects of data augmentation and the additional class for irrelevant questions, and to assess the models' consistency with human annotators. 



Table~\ref{tab:cls-results-aug} shows the BT classification results on the original DASQBT dataset and the effects of data augmentation. As outlined in the Methodology section, we expanded the training dataset by adding generated questions and BT labels using GPT-4o. Upon conducting data augmentation, we observe that the effects of data augmentation on this dataset are negative. This outcome is expected and explainable; given the inherent discrepancies between the original questions and those generated by GPT, we shifted the distribution of the training examples, leading to a slight decrease in F1-score from 0.836 to 0.825. Nevertheless, the high level of classification performance demonstrates its robustness in accurately identifying cognitive complexity levels of questions.



\begin{table}[hbtp]
  \scriptsize
  \centering
  \caption{BT Classification Results on \textbf{DASQBT} (Improvements with data augmentation in parentheses)}
  \label{tab:cls-results-aug}
  \begin{tabular}{@{}crrrrr@{}}
    \toprule
    \textbf{Class} & \textbf{Precision} & \textbf{Recall} & \textbf{F1-Score} & \textbf{Support} \\ \midrule
    Knowledge     & 0.824 (\phantom{\minus}0.054) & 0.875 (\minus 0.016) & 0.848 (\phantom{\minus}0.022) & 64 \\ 
    Comprehension & 0.798 (\minus 0.033) & 0.919 (\phantom{\minus}0.059) & 0.854 (\phantom{\minus}0.008) & 172 \\
    Application   & 0.783 (\minus 0.032) & 0.600 (\minus0.133) & 0.679 (\minus0.093) & 60 \\
    Analysis      & 0.912 (\minus 0.068) & 0.812 (\phantom{\minus}0.046) & 0.860 (\phantom{\minus}0.000) & 64 \\
    Evaluation    & 0.851 (\phantom{\minus}0.061) & 0.829 (\minus0.013) & 0.840 (\phantom{\minus}0.025) & 76 \\
    Synthesis     & 0.855 (\minus 0.027) & 0.768 (\minus0.102) & 0.809 (\minus0.067) & 69 \\ \midrule
    Macro avg     & 0.837 (\minus 0.008) & 0.801 (\minus 0.026) & 0.815 (\minus 0.017) & 505\\
    Weighted avg  & 0.830 (\minus 0.011) & 0.828 (\minus 0.008) & 0.825 (\minus 0.011) & 505 \\ 
    \bottomrule
  \end{tabular}
\end{table}

\begin{table}[hbtp]
  \scriptsize
  \centering
  \caption{BT Classification Results on \textbf{DASQBT with Irrel. Class} (Improvements with data augmentation in parentheses)}
  \label{tab:cls-results-aug-irrel}
  \begin{tabular}{@{}crrrrr@{}}
    \toprule
    \textbf{Class} & \textbf{Precision} & \textbf{Recall} & \textbf{F1-Score} & \textbf{Support} \\ \midrule
    Knowledge              & 0.747 (\minus0.074) & 0.875 (\phantom{\minus}0.016) & 0.806 (\minus0.034) & 64 \\ 
    Comprehension          & 0.809 (\minus0.006) & 0.860 (\minus0.064) & 0.834 (\minus0.032) & 172 \\
    Application            & 0.745 (\minus0.064) & 0.683 (\phantom{\minus}0.050) & 0.713 (\phantom{\minus}0.003) & 60 \\
    Analysis               & 0.945 (\phantom{\minus}0.086) & 0.812 (\minus0.047) & 0.874 (\phantom{\minus}0.015) & 64 \\
    Evaluation             & 0.905 (\phantom{\minus}0.023) & 0.750 (\minus0.039) & 0.820 (\minus0.013) & 76 \\
    Synthesis              & 0.811 (\phantom{\minus}0.008) & 0.870 (\phantom{\minus}0.102) & 0.839 (\phantom{\minus}0.054) & 69 \\ 
    Irrelevant    & 0.970 (\phantom{\minus}0.001) & 0.970 (\phantom{\minus}0.020) & 0.970 (\phantom{\minus}0.010) & 100 \\ \midrule
    Macro avg     & 0.847 (\minus0.004) & 0.832 (\phantom{\minus}0.006) & 0.837 (\phantom{\minus}0.001) & 605\\
    Weighted avg  & 0.849 (\minus0.003) & 0.845 (\minus0.006)& 0.845 (\minus0.004) & 605 \\ 
    \bottomrule
  \end{tabular}
\end{table}

Table~\ref{tab:cls-results-aug-irrel} shows the BT classification results with the additional \textit{Irrelevant} class. In this experiment, we identified and added 500 irrelevant questions to the dataset and properly split the dataset into training and testing sets. The results demonstrate that this method has improved the model's F1-score in predicting BT levels to 0.845, an increase of 2.0\%. The model also shows a high precision of 0.970 in identifying irrelevant questions, with the F1-score ranging from 0.713 to 0.874 across the BT levels. Notably, the model's ability to detect application-level questions has significantly improved in the second training phase, achieving an F1-score of 0.713. We observe a continued negative impact of data augmentation in this experiment, which is less pronounced than in the previous experiment.

\begin{figure}[ht]
  \centering
  \begin{subfigure}[b]{0.49\linewidth}
    \centering
    \includegraphics[width=\linewidth]{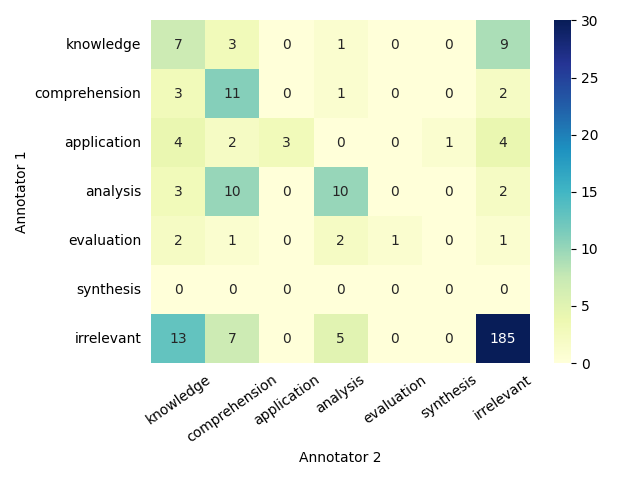}
    \caption{Annotator 1 vs. Annotator 2}
  \end{subfigure}
  \hfill
  \begin{subfigure}[b]{0.49\linewidth}
    \centering
    \includegraphics[width=\linewidth]{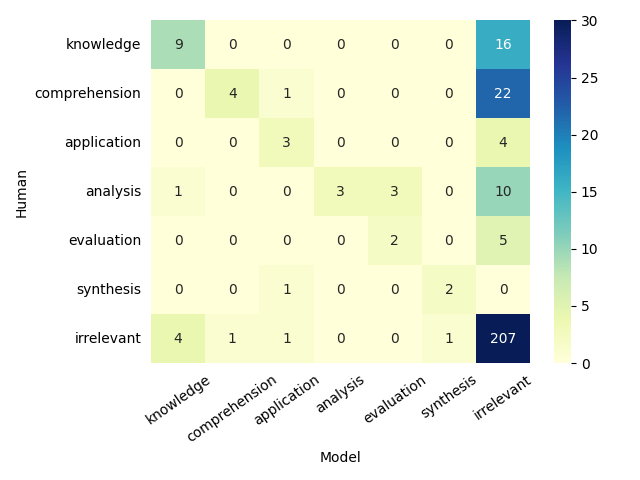}
    \caption{Human vs. Model}
  \end{subfigure}
  \caption{Confusion Matrix of Human Annotators and Model Prediction (Human labels are aggregated and agreed upon by 3 annotators)}
  \label{fig:confusion_matrix}
\end{figure}

The test on the human-annotated dataset evaluates the model's performance on the extracted YouTube question dataset. The results, as shown in Table~\ref{tab:cls-results-aug-irrel-human}, demonstrate a competitive weighted average F1-score of 0.721. It is clear that the effect of data augmentation is significant in this experiment, as the model's performance has improved considerably; we observe 16.7\% increase in macro average precision and 7.8\% increase in weighted average precision. The accompanying Confusion Matrix in Figure~\ref{fig:confusion_matrix} illustrates the agreement of annotations among human labelers and the model, highlighting the model's tendency to be more stringent in identifying relevant questions. We also observe relatively low recall scores across all BT categories except the \textit{Irrelevant} class. This is also due to the model's tendency mentioned above.

 \begin{table}[htbp]
  \scriptsize
\centering
  \caption{BT Classification Results on \textbf{300 Human-Labeled Samples} (Improvements with data augmentation in parentheses)}
 \label{tab:cls-results-aug-irrel-human}
\centering
  \begin{tabular}{@{}crrrrr@{}}
    \toprule
    \textbf{Accuracy} & \textbf{Precision} & \textbf{Recall} & \textbf{F1-Score} & \textbf{Support} \\ \midrule
    Knowledge              & 0.643 (\phantom{\minus}0.028) & 0.360 (\phantom{\minus}0.040) & 0.462 (\phantom{\minus}0.041) & 25 \\ 
    Comprehension          & 0.800 (\phantom{\minus}0.514) & 0.148 (\phantom{\minus}0.074) & 0.250 (\phantom{\minus}0.132) & 27 \\
    Application            & 0.500 (\phantom{\minus}0.000) & 0.429 (\phantom{\minus}0.286) & 0.462 (\phantom{\minus}0.241) & 7 \\
    Analysis               & 1.000 (\phantom{\minus}0.167) & 0.176 (\minus0.118) & 0.300 (\minus0.135) & 17 \\
    Evaluation             & 0.400 (\phantom{\minus}0.200) & 0.286 (\phantom{\minus}0.143) & 0.333 (\phantom{\minus}0.166) & 7 \\
    Synthesis              & 0.667 (\phantom{\minus}0.238) & 0.667 (\minus0.333) & 0.667 (\phantom{\minus}0.067) & 3 \\ 
    Irrelevant             & 0.784 (\phantom{\minus}0.019) & 0.967 (\phantom{\minus}0.037) & 0.866 (\phantom{\minus}0.026) & 214 \\ \midrule
    Macro avg     & 0.685 (\phantom{\minus}0.167) & 0.433 (\minus0.018) & 0.477 (\phantom{\minus}0.077) & 300\\
  Weighted avg  & 0.769 (\phantom{\minus}0.078)  & 0.767 (\phantom{\minus}0.037) & 0.721 (\phantom{\minus}0.037) & 300 \\ 
    \bottomrule
  \end{tabular}
\end{table}

\subsection{Distribution of BT Cognitive Levels across Subjects}

\begin{figure}[hbtp]
    \centering
    \includegraphics[width=1\linewidth]{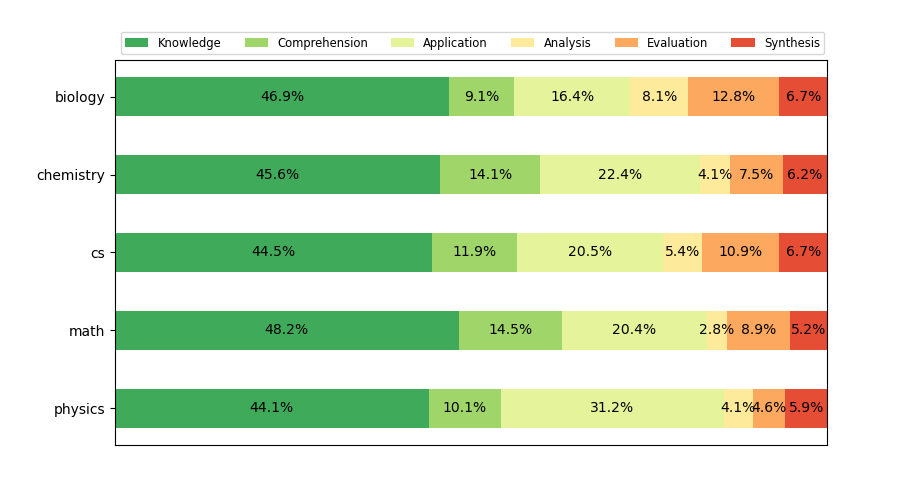}
    \caption{Question Type Distribution by Cognitive Complexity Across 5 Subjects}
    \label{fig:bt-cls-dist}
\end{figure}

In this section, we present a comparison of question types across the five subjects, as shown in Figure~\ref{fig:bt-cls-dist}. From the results, we observe that \textit{Knowledge} is the most prominent question type across all subjects, with percentages ranging from 44.1\% to 48.2\%. Notably, chemistry and math have higher proportions of \textit{Comprehension} questions compared to the other subjects. Physics, in particular, has the highest percentage of \textit{Application} questions at 31.2\%. Interestingly, the trained model tends to classify ``how to'' questions as \textit{Application} questions. Overall, higher-level questions (i.e., \textit{Analysis, Evaluation, Synthesis}) are less common, ranging from 14.6\% (physics) to 27.6\% (biology).

\subsection{Distinct Verb Analysis}

We conducted a more in-depth analysis of the model's predictions by identifying the distinct verbs in each BT class. We present the comparison of the top 10 verbs for each BT class in three of the datasets we have used in this study (i.e., \textit{(Augmented) DASQBT, YouLeQD}) and the commonly recognized set of verbs in the field of education (i.e., \textit{Standard}).

\begin{table*}[htbp]
  \centering
  \caption{Bloom's Taxonomy Verbs Identified from Our Datasets}
  \label{tab:bt-verbs}
  \begin{tabular}{@{}lll@{}}
  \toprule
  \textbf{Cognitive Level} & \textbf{Sources} &\textbf{Verbs} \\
  \midrule
  \multirow{4}{*}{\textbf{Knowledge}} 
     & \textit{Standard} & define, identify, describe, recognize, tell, explain, recite, memorize, illustrate, quote\\
     & \textit{Initial DASQBT} & recall, list, state, organize, live, build, briefly, name, define, follow\\
     & \textit{Augmented DASQBT} & state, list, define, mean, learning, come, stand, underlie, control, reach\\
     & \textit{YouTube Questions} & mature, happened, animal, species, shield, hash, size, length, enzyme, prescribe\\
  \midrule
  \multirow{4}{*}{\textbf{Comprehension}} 
     & \textit{Standard} & summarize, interpret, classify, compare, contrast, infer, relate, extract, paraphrase, cite\\
     & \textit{Initial DASQBT} & adapt, summarize, mean, classify, measure, form, retain, benefit, explain, describe\\
     & \textit{Augmented DASQBT} & help, explain, related, pack, regulate, differ, describe, undergo, allow, make\\
     & \textit{YouTube Questions} & summarise, detail, explain, dilate, deceive, explain, summarize, reside, convey, retire\\
  \midrule
  \multirow{4}{*}{\textbf{Application}} 
     & \textit{Standard} & solve, change, relate, complete, use, sketch, teach, articulate, discover, transfer\\
     & \textit{Initial DASQBT} & calculate, derive, compute, encode, demonstrate, specify, construct, apply, deploy, sketch\\
     & \textit{Augmented DASQBT} & use, hit, write, given, compress, spam, drop, recommend, exert, round\\
     & \textit{YouTube Questions} & slave, bacteria, len, int, revolve, inputting, temp, return, assemble, stimulate\\
  \midrule
  \multirow{4}{*}{\textbf{Analysis}} 
     & \textit{Standard} & contrast, connect, relate, devise, correlate, illustrate, distill, conclude, categorize, take apart\\
     & \textit{Initial DASQBT} & differentiate, distinguish, inspect, contrast, differ, examine, emerge, analyze, maintain, play\\
     & \textit{Augmented DASQBT} & contrast, highlight, fix, submerge, mine, differentiate, boost, group, experience, bag\\
     & \textit{YouTube Questions} & comparing, differentiate, routing, modern, distinguish, submit, align, gather, coordinate, impress\\
  \midrule
  \multirow{4}{*}{\textbf{Evaluation}}
     & \textit{Standard} & criticize, reframe, judge, defend, appraise, value, prioritize, plan, grade, reframe\\
     & \textit{Initial DASQBT} & defend, justify, evaluate, exercise, invest, agree, think, decide, predict, place\\
     & \textit{Augmented DASQBT} & justify, evaluate, reason, compare, versus, support, control, detect, understand, alter\\
     & \textit{YouTube Questions} & guarantee, arouse, false, prioritize, justify, govern, dispute, accumulate, judge, assert\\
  \midrule
  \multirow{4}{*}{\textbf{Synthesis}} 
     & \textit{Standard} & design, modify, role-play, develop, rewrite, pivot, modify, collaborate, invent, write\\
     & \textit{Initial DASQBT} & devise, compose, suggest, combine, introduce, synthesize, deal, propose, develop, overcome\\
     & \textit{Augmented DASQBT} & expect, outline, need, investigate, combine, collect, incorporate, design, synthesize, preprocessing\\
     & \textit{YouTube Questions} & tune, displace, naked, suggest, transform, obsess, propose, enhance, emerge, interbreed\\
  \bottomrule
  \end{tabular}
\end{table*}

To identify the distinct verbs, we used \textit{Pointwise Mutual Information} (PMI) and measured the association between the verb words and the cognitive levels by estimating the probability of the verb words and the cognitive levels using statistical information. The calculation formula is as follows:
\begin{equation}
    PMI(w, c) = \log \frac{P(w, c)}{P(w)P(c)},
\end{equation}
where $w$ is a verb word, $c$ is a BT class, $P(w, c)$ is the probability of the verb word and the class co-occurring, and $P(w)$ and $P(c)$ are the prior probabilities. The PMI value indicates the strength of the association between the verb word and cognitive level. We considered words with a collection frequency greater than 5 in order to avoid overfitting to low-frequency words. The results are shown in Table~\ref{tab:bt-verbs}.

Our analysis reveals that the list of verbs in the original DASQBT dataset closely corresponds with the commonly accepted list of words, namely \textit{Standard} in the table. This can be attributed to the curation of the DASQBT dataset, where the questions possess a strong association with BT level through the use of highly indicative words. Furthermore, the augmented DASQBT dataset also demonstrates a strong alignment with the standard list, also showing a diverse range of words from the GPT-generated questions. The verbs identified in the YouTube questions are more diverse and less indicative of cognitive levels compared to the DASQBT datasets. In general, This result is likely due to the informal nature of the comments, which may contain more colloquial language and less formal vocabulary.

The BT categories of \textit{Knowledge} and \textit{Application} exhibit a notable lack of coherence. We observe that the PMI values for these categories are significantly lower (1.925 compared to 3.782 in other categories), indicating a weak association between the verbs and the respective BT categories. We conjecture that this result may be attributed to the nature of the learner-posed questions. It is unlikely for YouTube learners to inquire about the definition of a concept or its application in problem-solving. Instead, their inquiries may involve seeking clarification or further explanation of a concept, which may not align directly with the two BT categories. Additionally, we note that certain words in the \textit{Application} category (such as len, int, temp, return, and head) are derived from programming language keywords. This indicates that the code snippets with the questions are classified as \textit{Application} questions, with certain words tagged as verbs.

\subsection{Cognitive Complexity and Engagement Rate}

\begin{figure}[hbtp]
  \centering
  \subfloat[Interaction Rate vs. BT Level]{{\includegraphics[width=0.5\linewidth]{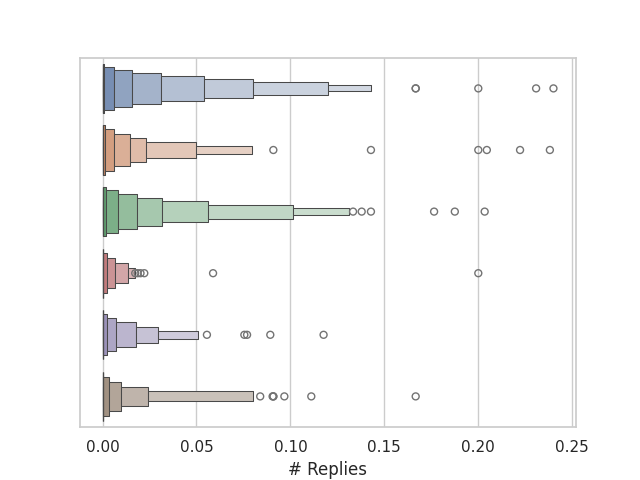} }}%
  \subfloat[Popularity vs. BT Level]{{\includegraphics[width=0.5\linewidth]{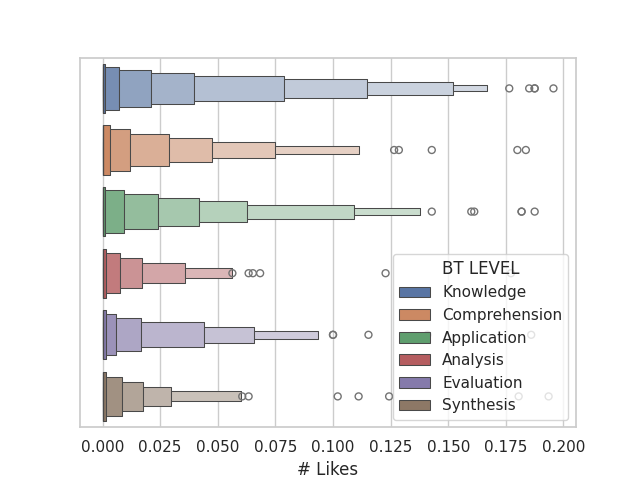} }}%
  \caption{Popularity and Interaction Rate vs. Bloom's Taxonomy on Cognitive Level}%
  \label{fig:pop-interaction}%
\end{figure}

Figure~\ref{fig:pop-interaction} shows the relationship between the BT cognitive levels and the interaction rate (\# of replies) and popularity (\# of votes/likes) of the questions. The rates are normalized by the number of views of the video. In common, \textit{Knowledge} and \textit{Application} show the highest counts, and \textit{Analysis} shows the least. We see an inverse relationship between popularity and BT cognitive levels. However, it is worth noting the presence of a positive correlation between interaction rate and the higher cognitive levels, specifically from \textit{Analysis} to \textit{Synthesis}.

\section{Conclusion and future work}

This study demonstrates the effectiveness of using a transformer-based model to automatically distinguish the cognitive complexity of questions based on Bloom's Taxonomy levels. Through the analysis of 57,242 learner-posed questions from YouTube educational video comments, we found that most questions fall under the lowest cognitive level, Knowledge. We also observed a complex relationship between question popularity and BT levels, as well as slight differences in cognitive levels across subjects. While there are limitations to this study, such as the small sample size of human annotation, our findings offer valuable insights into how students engage with online educational videos and the potential of GenAI models in the field of questioning.

\FloatBarrier
\bibliographystyle{IEEEtran}
\bibliography{references}

\begin{thebibliography}{10}
\providecommand{\url}[1]{#1}
\csname url@samestyle\endcsname
\providecommand{\newblock}{\relax}
\providecommand{\bibinfo}[2]{#2}
\providecommand{\BIBentrySTDinterwordspacing}{\spaceskip=0pt\relax}
\providecommand{\BIBentryALTinterwordstretchfactor}{4}
\providecommand{\BIBentryALTinterwordspacing}{\spaceskip=\fontdimen2\font plus
\BIBentryALTinterwordstretchfactor\fontdimen3\font minus \fontdimen4\font\relax}
\providecommand{\BIBforeignlanguage}[2]{{%
\expandafter\ifx\csname l@#1\endcsname\relax
\typeout{** WARNING: IEEEtran.bst: No hyphenation pattern has been}%
\typeout{** loaded for the language `#1'. Using the pattern for}%
\typeout{** the default language instead.}%
\else
\language=\csname l@#1\endcsname
\fi
#2}}
\providecommand{\BIBdecl}{\relax}
\BIBdecl

\bibitem{king1990enhancing}
A.~King, ``Enhancing peer interaction and learning in the classroom through reciprocal questioning,'' \emph{American educational research journal}, vol.~27, no.~4, pp. 664--687, 1990.

\bibitem{chin2002student}
C.~Chin and D.~E. Brown, ``Student-generated questions: A meaningful aspect of learning in science,'' \emph{International Journal of Science Education}, vol.~24, no.~5, pp. 521--549, 2002.

\bibitem{rosenshine1996teaching}
B.~Rosenshine, C.~Meister, and S.~Chapman, ``Teaching students to generate questions: A review of the intervention studies,'' \emph{Review of educational research}, vol.~66, no.~2, pp. 181--221, 1996.

\bibitem{borah2024improved}
A.~R. Borah, T.~Nischith, and S.~Gupta, ``Improved learning based on genai,'' in \emph{2024 2nd International Conference on Intelligent Data Communication Technologies and Internet of Things (IDCIoT)}.\hskip 1em plus 0.5em minus 0.4em\relax IEEE, 2024, pp. 1527--1532.

\bibitem{nyaaba2024generative}
M.~Nyaaba, L.~Shi, M.~Nabang, X.~Zhai, P.~Kyeremeh, S.~A. Ayoberd, and B.~N. Akanzire, ``Generative ai as a learning buddy and teaching assistant: Pre-service teachers' uses and attitudes,'' \emph{arXiv preprint arXiv:2407.11983}, 2024.

\bibitem{luo2024does}
J.~Luo, ``How does genai affect trust in teacher-student relationships? insights from students’ assessment experiences,'' \emph{Teaching in Higher Education}, pp. 1--16, 2024.

\bibitem{grainger2018process}
R.~Grainger, E.~Osborne, W.~Dai, and D.~Kenwright, ``The process of developing a rubric to assess the cognitive complexity of student-generated multiple choice questions in medical education,'' \emph{The Asia Pacific Scholar}, vol.~3, no.~2, p.~19, 2018.

\bibitem{teplitski2018student}
M.~Teplitski, T.~Irani, C.~J. Krediet, M.~Di~Cesare, and M.~Marvasi, ``Student-generated pre-exam questions is an effective tool for participatory learning: A case study from ecology of waterborne pathogens course,'' \emph{Journal of Food Science Education}, vol.~17, no.~3, pp. 76--84, 2018.

\bibitem{irons2014coaching}
N.~A. Irons, ``Coaching for questioning: A study on the impact of questioning,'' \emph{Unpublished Capstone Action Research Project). Fielding Graduate University, Santa Barbara, CA}, 2014.

\bibitem{kruse2009gagne}
K.~Kruse, ``Gagne's nine events of instruction: An introduction,'' \emph{Retrieved the}, vol.~10, 2009.

\bibitem{bergsteiner2010kolb}
H.~Bergsteiner, G.~C. Avery, and R.~Neumann, ``Kolb's experiential learning model: critique from a modelling perspective,'' \emph{Studies in continuing education}, vol.~32, no.~1, pp. 29--46, 2010.

\bibitem{merrill2002first}
M.~D. Merrill, ``First principles of instruction,'' \emph{Educational technology research and development}, vol.~50, pp. 43--59, 2002.

\bibitem{kuhail2023interacting}
M.~A. Kuhail, N.~Alturki, S.~Alramlawi, and K.~Alhejori, ``Interacting with educational chatbots: A systematic review,'' \emph{Education and Information Technologies}, vol.~28, no.~1, pp. 973--1018, 2023.

\bibitem{lee2022developing}
D.~Lee and S.~Yeo, ``Developing an ai-based chatbot for practicing responsive teaching in mathematics,'' \emph{Computers \& Education}, vol. 191, p. 104646, 2022.

\bibitem{tran2023generating}
A.~Tran, K.~Angelikas, E.~Rama, C.~Okechukwu, D.~H. Smith, and S.~MacNeil, ``Generating multiple choice questions for computing courses using large language models,'' in \emph{2023 IEEE Frontiers in Education Conference (FIE)}.\hskip 1em plus 0.5em minus 0.4em\relax IEEE, 2023, pp. 1--8.

\bibitem{gnanasekaran2021automatic}
D.~Gnanasekaran, R.~Kothandaraman, and K.~Kaliyan, ``An automatic question generation system using rule-based approach in bloom’s taxonomy,'' \emph{Recent Advances in Computer Science and Communications (Formerly: Recent Patents on Computer Science)}, vol.~14, no.~5, pp. 1477--1487, 2021.

\bibitem{van2022evaluating}
R.~Van~Campenhout, M.~Hubertz, and B.~G. Johnson, ``Evaluating ai-generated questions: a mixed-methods analysis using question data and student perceptions,'' in \emph{International Conference on Artificial Intelligence in Education}.\hskip 1em plus 0.5em minus 0.4em\relax Springer, 2022, pp. 344--353.

\bibitem{bhat2022towards}
S.~Bhat, H.~A. Nguyen, S.~Moore, J.~C. Stamper, M.~Sakr, and E.~Nyberg, ``Towards automated generation and evaluation of questions in educational domains.'' in \emph{EDM}, 2022.

\bibitem{jamleh2021evaluation}
A.~Jamleh, M.~Nassar, H.~Alissa, and A.~Alfadley, ``Evaluation of youtube videos for patients’ education on periradicular surgery,'' \emph{Plos one}, vol.~16, no.~12, p. e0261309, 2021.

\bibitem{clifton2011can}
A.~Clifton and C.~Mann, ``Can youtube enhance student nurse learning?'' \emph{Nurse education today}, vol.~31, no.~4, pp. 311--313, 2011.

\bibitem{wahyuni2021use}
A.~Wahyuni, A.~R. Utami, and E.~Education, ``The use of youtube video in encouraging speaking skill,'' \emph{Pustakailmu. id}, vol.~7, no.~3, pp. 1--9, 2021.

\bibitem{fadhil2020investigating}
N.~Fadhil~Abbas and T.~Ali~Qassim, ``Investigating the effectiveness of youtube as a learning tool among efl students at baghdad university,'' \emph{Arab World English Journal (AWEJ) Special Issue on CALL}, no.~6, 2020.

\bibitem{dubovi2020empirical}
I.~Dubovi and I.~Tabak, ``An empirical analysis of knowledge co-construction in youtube comments,'' \emph{Computers \& Education}, vol. 156, p. 103939, 2020.

\bibitem{meyers2014comment}
E.~M. Meyers, ``A comment on learning: Media literacy practices in youtube,'' \emph{international Journal of Learning and Media}, vol.~4, no. 3-4, 2014.

\bibitem{lee2017making}
C.~S. Lee, H.~Osop, D.~H.-L. Goh, and G.~Kelni, ``Making sense of comments on youtube educational videos: A self-directed learning perspective,'' \emph{Online information review}, vol.~41, no.~5, pp. 611--625, 2017.

\bibitem{wang2023sight}
R.~E. Wang, P.~Wirawarn, N.~Goodman, and D.~Demszky, ``Sight: A large annotated dataset on student insights gathered from higher education transcripts,'' \emph{arXiv preprint arXiv:2306.09343}, 2023.

\bibitem{papinczak2012using}
T.~Papinczak, R.~Peterson, A.~S. Babri, K.~Ward, V.~Kippers, and D.~Wilkinson, ``Using student-generated questions for student-centred assessment,'' \emph{Assessment \& Evaluation in Higher Education}, vol.~37, no.~4, pp. 439--452, 2012.

\bibitem{dillon2004questioning}
J.~T. Dillon, \emph{Questioning and teaching: A manual of practice}.\hskip 1em plus 0.5em minus 0.4em\relax Wipf and Stock Publishers, 2004.

\bibitem{gatt2018survey}
A.~Gatt and E.~Krahmer, ``Survey of the state of the art in natural language generation: Core tasks, applications and evaluation,'' \emph{Journal of Artificial Intelligence Research}, vol.~61, pp. 65--170, 2018.

\bibitem{amidei2018evaluation}
J.~Amidei, P.~Piwek, and A.~Willis, ``Evaluation methodologies in automatic question generation 2013-2018,'' in \emph{Proceedings of the 11th International Conference on Natural Language Generation}, 2018, pp. 307--317.

\bibitem{kurdi2020systematic}
G.~Kurdi, J.~Leo, B.~Parsia, U.~Sattler, and S.~Al-Emari, ``A systematic review of automatic question generation for educational purposes,'' \emph{International Journal of Artificial Intelligence in Education}, vol.~30, pp. 121--204, 2020.

\bibitem{mulla2023automatic}
N.~Mulla and P.~Gharpure, ``Automatic question generation: a review of methodologies, datasets, evaluation metrics, and applications,'' \emph{Progress in Artificial Intelligence}, vol.~12, no.~1, pp. 1--32, 2023.

\bibitem{haladyna2021using}
T.~M. Haladyna and M.~C. Rodriguez, ``Using full-information item analysis to improve item quality,'' \emph{Educational Assessment}, vol.~26, no.~3, pp. 198--211, 2021.

\bibitem{gierl2016evaluating}
M.~J. Gierl, H.~Lai, D.~Pugh, C.~Touchie, A.-P. Boulais, and A.~De~Champlain, ``Evaluating the psychometric characteristics of generated multiple-choice test items,'' \emph{Applied Measurement in Education}, vol.~29, no.~3, pp. 196--210, 2016.

\bibitem{wind2019exploring}
S.~A. Wind, M.~Alemdar, J.~A. Lingle, R.~Moore, and A.~Asilkalkan, ``Exploring student understanding of the engineering design process using distractor analysis,'' \emph{International Journal of STEM Education}, vol.~6, pp. 1--18, 2019.

\bibitem{vsmida2024developing}
D.~{\v{S}}mida, E.~{\v{C}}ipkov{\'a}, and M.~Fuchs, ``Developing the test of inquiry skills: measuring the level of inquiry skills among pupils in slovakia,'' \emph{International Journal of Science Education}, vol.~46, no.~1, pp. 73--108, 2024.

\bibitem{rezigalla2024item}
A.~A. Rezigalla, A.~M. E. S.~A. Eleragi, A.~B. Elhussein, J.~Alfaifi, M.~A. ALGhamdi, A.~Y. Al~Ameer, A.~I.~O. Yahia, O.~A. Mohammed, and M.~I.~E. Adam, ``Item analysis: the impact of distractor efficiency on the difficulty index and discrimination power of multiple-choice items,'' \emph{BMC Medical Education}, vol.~24, no.~1, p. 445, 2024.

\bibitem{bloom1956handbook}
B.~S. Bloom, M.~D. Engelhart, E.~Furst, W.~H. Hill, and D.~R. Krathwohl, ``Handbook i: cognitive domain,'' \emph{New York: David McKay}, pp. 483--498, 1956.

\bibitem{roth2007understanding}
D.~Roth, ``Understanding by design: A framework for effecting curricular development and assessment,'' \emph{CBE—Life Sciences Education}, vol.~6, no.~2, pp. 95--97, 2007.

\bibitem{marzano1998implementing}
R.~J. Marzano and J.~S. Kendall, ``Implementing standards-based education. student assessment series.'' 1998.

\bibitem{fink2013creating}
L.~D. Fink, \emph{Creating significant learning experiences: An integrated approach to designing college courses}.\hskip 1em plus 0.5em minus 0.4em\relax John Wiley \& Sons, 2013.

\bibitem{mohammed2020question}
M.~Mohammed and N.~Omar, ``Question classification based on bloom’s taxonomy cognitive domain using modified tf-idf and word2vec,'' \emph{PloS one}, vol.~15, no.~3, p. e0230442, 2020.

\bibitem{ahmed2024classification}
H.~M. Ahmed and S.~E. Sorour, ``Classification-driven intelligent system for automated evaluation of higher education exam paper quality,'' \emph{Education and Information Technologies}, pp. 1--27, 2024.

\bibitem{das2020identification}
S.~Das, S.~K.~D. Mandal, and A.~Basu, ``Identification of cognitive learning complexity of assessment questions using multi-class text classification,'' \emph{Contemporary Educational Technology}, vol.~12, no.~2, p. ep275, 2020.

\bibitem{fan2023bibliometric}
L.~Fan, L.~Li, Z.~Ma, S.~Lee, H.~Yu, and L.~Hemphill, ``A bibliometric review of large language models research from 2017 to 2023,'' \emph{arXiv preprint arXiv:2304.02020}, 2023.

\bibitem{aguda2024large}
T.~Aguda, S.~Siddagangappa, E.~Kochkina, S.~Kaur, D.~Wang, C.~Smiley, and S.~Shah, ``Large language models as financial data annotators: A study on effectiveness and efficiency,'' \emph{arXiv preprint arXiv:2403.18152}, 2024.

\bibitem{he2023annollm}
X.~He, Z.~Lin, Y.~Gong, A.~Jin, H.~Zhang, C.~Lin, J.~Jiao, S.~M. Yiu, N.~Duan, W.~Chen \emph{et~al.}, ``Annollm: Making large language models to be better crowdsourced annotators,'' \emph{arXiv preprint arXiv:2303.16854}, 2023.

\bibitem{chen2024large}
R.~Chen, C.~Qin, W.~Jiang, and D.~Choi, ``Is a large language model a good annotator for event extraction?'' in \emph{Proceedings of the AAAI Conference on Artificial Intelligence}, vol.~38, no.~16, 2024, pp. 17\,772--17\,780.

\bibitem{chen2024qgen}
M.-S. Chen and A.-Z. Yen, ``E-qgen: Educational lecture abstract-based question generation system,'' \emph{arXiv preprint arXiv:2404.13547}, 2024.

\bibitem{liu2019roberta}
Y.~Liu, M.~Ott, N.~Goyal, J.~Du, M.~Joshi, D.~Chen, O.~Levy, M.~Lewis, L.~Zettlemoyer, and V.~Stoyanov, ``Roberta: A robustly optimized bert pretraining approach,'' \emph{arXiv preprint arXiv:1907.11692}, 2019.

\bibitem{Khan_2021}
\BIBentryALTinterwordspacing
S.~Khan, ``Questions vs statements classification,'' May 2021. [Online]. Available: \url{https://www.kaggle.com/datasets/shahrukhkhan/questions-vs-statementsclassificationdataset}
\BIBentrySTDinterwordspacing

\bibitem{rajpurkar2016squad}
P.~Rajpurkar, J.~Zhang, K.~Lopyrev, and P.~Liang, ``Squad: 100,000+ questions for machine comprehension of text,'' \emph{arXiv preprint arXiv:1606.05250}, 2016.

\bibitem{leech2013spaadia}
G.~Leech and M.~Weisser, ``The spaadia annotation scheme,'' 2013.

\bibitem{hinton2015distilling}
G.~Hinton, ``Distilling the knowledge in a neural network,'' \emph{arXiv preprint arXiv:1503.02531}, 2015.

\bibitem{zhou2021rethinking}
H.~Zhou and L.~Song, ``Rethinking soft labels for knowledge distillation: A bias--variance tradeoff perspective,'' in \emph{Proceedings of International Conference on Learning Representations (ICLR)}, 2021.

\bibitem{hendrycks2022baseline}
D.~Hendrycks and K.~Gimpel, ``A baseline for detecting misclassified and out-of-distribution examples in neural networks,'' in \emph{International Conference on Learning Representations}, 2022.

\end{thebibliography}


\clearpage
\newpage
\appendices
\section{Confusing Statements in BT Questions Classification\label{appx:low-conf-sample}}

The log probabilities are used to discern challenging instances for the GPT-4o model. Table~\ref{tab:low-conf-samples} shows some examples with low confidence scores. These confusion statements can largely be attributed to the following anomalies:

\begin{enumerate}
  \item Removal of the ending question marks.
  \item Incomplete sentences.
  \item Use of wh-words in non-interrogative sentences.
  \item Mix of interrogative and non-interrogative sentences in one statement.
\end{enumerate}

\begin{table}[htbp]
  \centering
  \caption{Confusing Statements with Low Confidence Scores from GPT-4o}
  \label{tab:low-conf-samples}
  \begin{tabular}{p{0.7\linewidth}r}
  \toprule
  \textbf{Text} & \textbf{Logprobs} \\
  \midrule
  Houston was founded in 1836 on land near the banks of Buffalo Bayou (now known as Allen's Landing) and incorporated as a city on June 5, 1837. When was Houston founded & -0.6935 \\
  Portuguese universities have existed since 1290. Since when have Portuguese universities existed & -5.7273 \\
  Understanding who's work was the tradition meaning of Cubism formed on & -0.5759 \\
  The optimum format for a broadcast depends upon the type of videographic recording medium used and the image's characteristics. The type of videographic recording medium used and the image's characteristics determine what? & -1.138\\
  \bottomrule
  \end{tabular}
\end{table}

\section{Details of Prompt Engineering}
\label{appx:prompts}

\subsection{Prompts in Question Classification for Knowledge Distillation}

In the question (\textit{Interrogative vs. Declarative}) classification model, we use the following prompt:\\[.7em]
\textbf{System:} \textit{You will be asked to classify the following sentence into interrogative (question) or non-interrogative (statement). Use the labels '0' for non-interrogative and '1' for interrogative.}

\textbf{User:} \textit{In 1952, Thomas Watson, Sr. In what year did IBM open its first office in Poughkeepsie}

\textbf{Assistant} \textit{1}

\textbf{User:} \textit{What often lacks in software developed when its released that can eventually lead to errors?}

\textbf{Assistant} \textit{1}

\textbf{User:} \textit{there's nothing really that gets in that early}

\textbf{Assistant} \textit{0}

\textbf{User:} \textit{The President of BYU, currently Kevin J Worthen, reports to the Board, through the Commissioner of Education.}

\textbf{Assistant} \textit{0}

\textbf{User:} \textbf{\{\textit{Test\_statement}\}}

\subsection{Prompts in BT Classification for Data Augmentation}

In training the BT classification model, we use the following prompt to augment training examples:\\[.7em]
\textbf{System:}  \textit{Bloom's Taxonomy categorizes cognitive levels into six levels, each described by a specific action verb:}

{\itshape
\begin{itemize} 
    \item Knowledge: define, match, recall, state, list, label
    \item Comprehension: discuss, review, paraphrase, describe, explain
    \item Application: apply, demonstrate, illustrate, solve, use
    \item Analysis: analyze, compare, contrast, differentiate, distinguish
    \item Evaluation: argue, conclude, critique, evaluate, justify, verify
    \item Synthesis: create, design, develop, formulate, organize, plan
\end{itemize}
}
\textit{The following are examples of questions that correspond to one of Bloom's Cognitive Levels. Please write a question that aligns with the given Bloom's Cognitive Level and pertains to a STEM topic.}

\textbf{User:} \textit{Generate a question that belongs to \textbf{\{BT\_level\}} level. }

\textbf{System:} \textit{\textbf{\{Random Sample Question in the corresponding BT\_level\}}.}

[Ends the few-shot examples above]

\textbf{User:}  \textit{Generate a \textbf{\{BT\_level\}} question on \textbf{\{Topic\}} in \textbf{\{Subject\}}}

\section{ Question Type Review}

\label{appx:question-type-review}

This section offers a comprehensive overview of the common question types present in the dataset. It provides an in-depth description of the main categories, covering the characteristics and purposes that define most of the questions in the dataset. The review aims to offer a clearer understanding of the nature of the questions in the dataset.

\subsection{Question Type 1: Questions Asking for Study Resources}

\textbf{Description}: Questions where learners ask for study materials, such as problem sets, apps, or answers from other learners or instructors.

\textbf{Examples}:
\begin{itemize}  
    \item “Could you guys please upload the problem sets for 5.111 on the OCW web page?”
    \item “Anyone knows some way to get the solutions for the assignments?”
\end{itemize}

\textbf{Criteria}:
\begin{itemize}  
    \item Clear request for study materials.
    \item Directed towards the instructor or fellow learners.
    \item Typically includes references to specific resources (problem sets, apps, textbooks).
\end{itemize}

\subsection{Question Type 2: Nonsensical and Meaningless Questions}

\textbf{Description}: Short, unclear questions that are difficult to understand or seem unrelated to the context.

\textbf{Examples}:
\begin{itemize} 
    \item “Big guns?”
    \item “How gonads go?”
\end{itemize}

\textbf{Criteria}:
\begin{itemize} 
    \item Lack of context or clear meaning.
    \item Too short to convey any substantial information.
    \item Unrelated to the topic being discussed.
\end{itemize}

\subsection{Question Type 3: Rhetorical questions}

\textbf{Description}: These comments look like questions but are often statements or compliments in disguise. They don’t seek an actual answer.

\textbf{Examples}:
\begin{itemize} 
    \item “I cannot be the only one in 2018 watching this for biology?”
    \item “How is this guy a professor in more than 10 subjects and a god in debating?”
\end{itemize}

\textbf{Criteria}:
\begin{itemize} 
    \item Structured as a question but does not seek information.
    \item Contains personal opinions or compliments.
    \item Expresses admiration or personal reflection rather than inquiry.
\end{itemize}

\subsection{Question Type 4: Interaction-based Questions}

\textbf{Description}: Questions aimed at interacting with other learners, typically discussing shared experiences or feelings about the video.

\textbf{Examples}: 
\begin{itemize} 
    \item “Who else here because of biology teacher?”
    \item “Is it just me or is this video really satisfying?”
\end{itemize}

\textbf{Criteria}:
\begin{itemize} 
    \item Engages the community of learners.
    \item Often includes informal, casual language.
    \item Seeks connection or shared experiences rather than factual answers.
\end{itemize}

\subsection{Question Type 5: Questions Regarding Lecture Content}

\textbf{Description}: These questions specifically address the lecture material, often prefacing the actual question with a statement.

\textbf{Examples}:
\begin{itemize}
    \item “So Hank, I have a question. (But first, a statement). Because of the domestication of less-aggressive, more prone-to-human-contact wolves, dogs have evolved into their own species. However, I have now seen Wolf/Dog hybrids being sold online. Are they creating pedigree lines with one set of parents or are they constantly breeding the same pairs?”
    \item “Can I ask a question? In the magnetic field, what does the ‘u’ mean in ‘uT’?”
    \item “Where did the A term come from at 4:22?”
\end{itemize}

\textbf{Criteria}:
\begin{itemize} 
    \item References lecture content directly.
    \item Often contains a background statement before asking the question.
    \item Focuses on clarifying or expanding on information provided in the lecture.
\end{itemize}

\end{document}